\documentclass{article}

\usepackage{times}

\usepackage{amssymb,amsmath}
\usepackage{epsfig,graphicx}
\usepackage{verbatim}
\usepackage{amsfonts}

\long\def\comment#1{}

\newcommand{\D}{\:\: |\:\:}
\newtheorem{examp}{Example}

\newenvironment{example}{\begin{examp}\rm}{\end{examp}}

\title{Structured Production System (extended abstract)}
\author{Yi Zhou}

\date{}

\begin{document}

\maketitle

\begin{abstract}
In this extended abstract, we propose Structured Production Systems
(SPS), which extend traditional production systems with well-formed
syntactic structures. Due to the richness of structures, structured
production systems significantly enhance the expressive power as
well as the flexibility of production systems, for instance, to
handle uncertainty. We show that different rule application
strategies can be reduced into the basic one by utilizing
structures. Also, many fundamental approaches in computer science,
including automata, grammar and logic, can be captured by structured
production systems.
\end{abstract}

\section{Introduction}

Production system is one of the most important approaches in AI.
Simply enough, a production system contains a set of production
rules of the form:
\begin{equation}\label{simple-production-rule}
a_1, \dots, a_n \to b,
\end{equation}
where $\{a_1,\dots, a_n\}$ is a set of preconditions called the
antecedent, and $b$ is an action or a postcondition called the
consequent. If the preconditions are satisfied by the current state
of the world, then the production rule can be triggered and applied,
and consequently, the action can be executed or the postcondition
can be obtained. Production systems are widely applied in many
application domains including expert systems
\cite{Brownston85,Hayes-Roth83,Jackson98}, action selection in
robotics \cite{Arkin98,Brooks85} and natural language processing
\cite{Allen95}.

Production system has many advantages. Firstly, although simple,
production system is computationally very powerful. Many production
system based computational models, e.g., Post canonical system, are
Turing complete \cite{Minsky67}. Secondly, production system is
highly modular. Last but not least, production rules are very
intuitive to be understood and used by human users.

Nevertheless, production system also has some critical
disadvantages. One of the main concerns is that it is not expressive
enough to handle sophisticated knowledge, e.g., uncertainty and
logic. Another concern is flexibility, that is, traditional
production systems only trigger and apply rules one-by-one, which is
not flexible enough to incorporate other rule application strategies
such as simultaneous rule application. Succinctness is also an
issue. In some cases, it might need too many production rules to
model an application domain.

Consider an application domain in automated solving intelligence
test questions, including sequencing number games. We need to
represent different forms of patterns such as cube and Fibonacci,
and their potential combinations. Also, we need to deal with
probabilities because for a given sequencing number problem, the way
to solve it could be a probabilistic distribution over different
patterns. For such a challenging task, we need a production system
that is not only efficient, but also flexible and expressive enough
to represent and reason about different kinds of sophisticated
knowledge.

To address these issues, we propose structured production systems.
Roughly speaking, a structured production system is a production
system augmented with a well-formed syntactic structure, which is a
rich framework to represent objects and knowledge in the application
domain.

The richness of syntactic structures brings a lot of benefits to
production systems. First of all, the syntactic structure can model
more sophisticated objects so that, both antecedents and consequents
in production systems can represent more sophisticated knowledge,
including uncertainty information and logic sentences. Secondly,
with the syntactic structure, one can flexibly apply rules, e.g., to
trigger one or many rules to be applied at the same time. Thirdly,
we show that structured production system serves as a general
framework for automated reasoning and modeling dynamics in the sense
that it can capture many existing approaches, including grammars,
automata, abstract rewriting systems, logic axiom systems and so on.

The rest of this extended abstract is organized as follows. Section
2 briefly reviews the basic notions and notations about syntactic
structures and assertions. Section 3 proposes structured production
systems that contain a syntactic structure and a set of ground and
schema rules. Section 4 shows that different rule application
strategies can be reduced into the basic one by utilizing
structures. Then, Section 5 shows that structured production systems
provide a general framework for modeling dynamics as it can capture
many important approaches in computer science and artificial
intelligence. Section 6 shows how to handle uncertainty in
structured production systems. Finally, Section 7 discusses related
issues and concludes this extended abstract.

\section{Structures, Terms and Assertions}

We assume the readers are familiar with some basic notions and
notations in set theory. A {\em syntactic structure} ({\em
structure} for short) is a triple $\langle \mathcal{I}, \mathcal{C},
\mathcal{O} \rangle$, where $\mathcal{I}$ is a class of {\em
individuals}, representing objects in an application domain;
$\mathcal{C}$ is a class of {\em concepts}, representing groups of
individuals that share something in common. Essentially, concepts
are sets in the sense that for each concept $C \in \mathcal{C}$, $C
\subseteq \mathcal{I}$; $\mathcal{O}$ is a class of {\em operators}
on individuals, representing interrelationships among individuals
and concepts in the application domain. Each operator is associated
with a domain of the form  $(C_1,\dots, C_n)$, representing all
possible values that the operator $O$ can operate on, where $C_i \in
\mathcal{C}, 1 \le i \le n$. Here, $n$ is called the arity of $O$.
For an $n$-ary tuple $(a_1, \dots, a_n)$ matching the domain of an
operator $O$, i.e., $a_i \in C_i, 1 \le i \le n$, $O$ maps it into a
new individual, denoted by $O(a_1, \dots, a_n)$. Concepts and
operators can be treated as individuals as well. In this sense, if
needed, we can have a concept that is a collection of concepts, a
concept that is a collection of operators and so on.

\begin{example}\label{door-scenario-1}
\begin{figure}
\centering
\includegraphics[width=5cm]{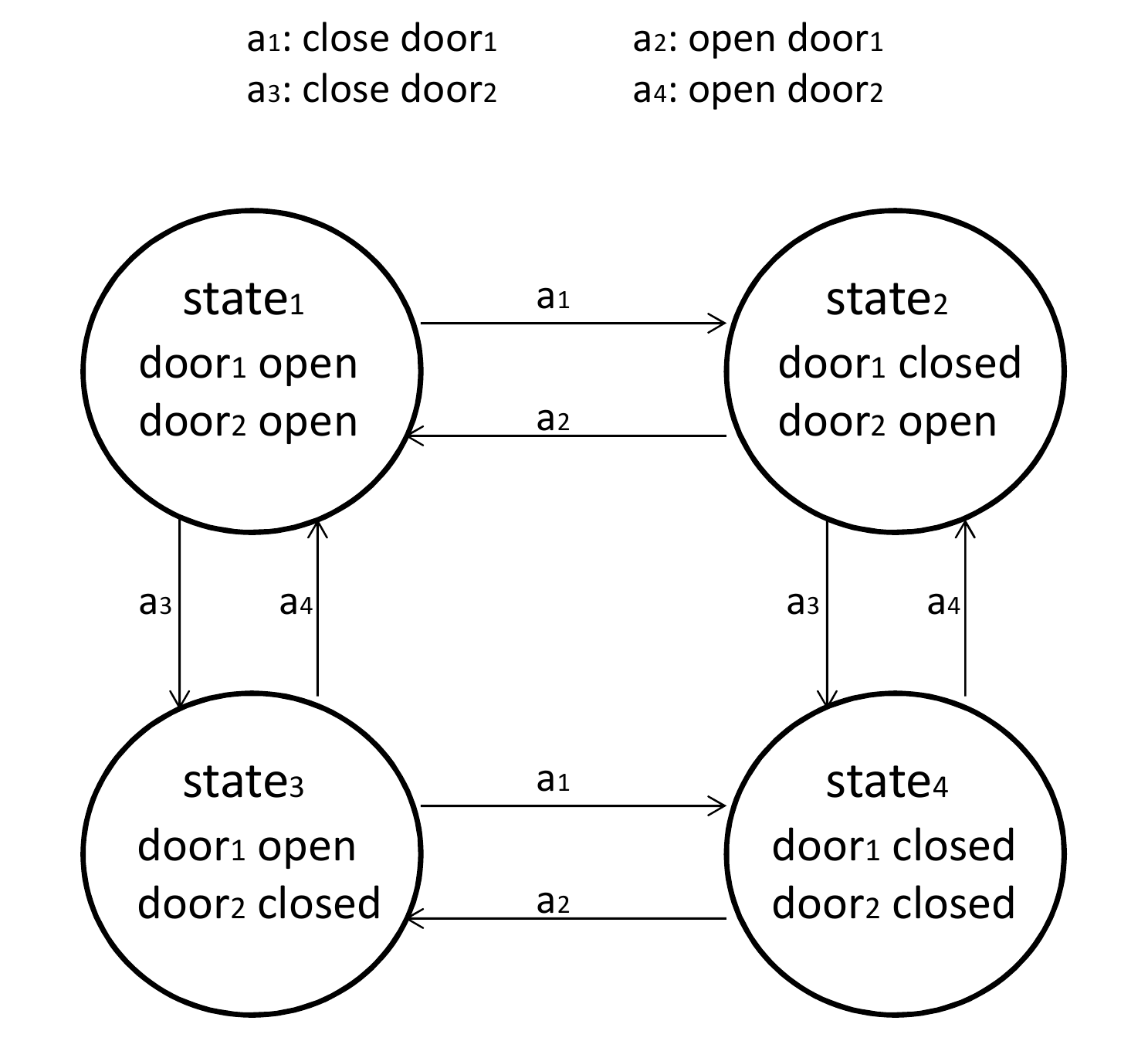}
\caption{Door control: a simple application domain}
\label{fig-example}
\end{figure}
Figure \ref{fig-example} depicts a simple application domain for
opening/closing two doors. To formalize this domain, one can use
state transition systems. There are four states in this scenario. At
$state_1$, both doors are open. If an action $a_1$ is successfully
executed to close $door_1$, then $state_1$ is transited into
$state_2$, in which $door_2$ is still open while $door_1$ is closed.
Nevertheless, state transition system has the {\em state explosion
problem} as there could be too many states to be exhausted. Suppose
that we generalize the door scenario into $n$ doors. Then, a state
transition system needs to use $2^n$ states, $2\times n$ actions and
$n \times 2^n$ transitions in order to model this domain.

Therefore, we need to use syntax for the sake of succinctness.
Suppose that we have $n$ doors. Each door $door_1, \dots, door_n$ is
an individual and they together form the concept $Door$. Each door
has two status, either $o$ (for ``open") or $c$ (for ``closed").
Then, $Status$ is an operator whose domain is $Door$ and whose value
can be either individual $o$ or individual $c$. For actions, there
are two action operators, namely $Open$ and $Close$, whose domains
are both $Door$. For a particular individual in $Door$, e.g.,
$door_1$, the action operator yields an action individual, e.g.,
$Open(door_1)$. We also introduce a specific operator $Do$ whose
domain is the concept of all actions and whose value can be either
$true$ or $false$.
\end{example}

{\em Terms} are defined recursively as follows:
\begin{itemize}
\item an individual is a term;
\item the result an operator $O$ operating on a tuple $(t_1, \dots,t_n)$ of terms that
matches the domain of $O$ is also a term.
\end{itemize}
Then, an {\em assertion} is of the form
\begin{equation}\label{form-assertion}
t_1=t_2,
\end{equation}
where $t_1$ and $t_2$ are two terms. In particular, if $t_2$ is the
individual ``$true$" for representing true statements, we omit it
and the associated equality symbol $=$ in the assertion for
simplicity. Also, terms and assertions can be considered as
individuals to be studied as well.

\begin{example}\label{door-scenario-2}[Example \ref{door-scenario-1} continued]
According to the definitions, $door_1$, $Status(door_1)$,
$Open(door_1)$, $Do(Close(door_1))$ are terms. $Status(door_1)=c$
and $Do(Close(door_1))=true$ are assertions, and the latter can be
simplified as $Do(Close(door_1))$.
\end{example}

\section{Rules and Structured Production Systems}

In this section, we present the formal definition of structured
production systems. First of all, we define production rules. A {\em
ground rule} is of the rule form (\ref{simple-production-rule})
except that the preconditions $a_1,\dots, a_n$ and the postcondition
$b$ are specified to be assertions defined in Section 2.

Other than ground rules, we also introduce schema rules. Similar to
concepts that group individuals, schema rules are used to group
ground rules. A {\em schema rule} contains two parts:
\begin{itemize}
\item a set of variable declarations of the form
\begin{equation}\label{variable-declaration}
x:C,
\end{equation}
where $x$ is a variable ranging over all individuals in $C$.
\item a rule part of the form
\begin{equation}\label{schema-rule}
a_1,\dots, a_n \to b,
\end{equation}
where $a_1,\dots, a_n$ and $b$ are assertions except that
individuals occurred in the rule could be replaced by variables
declared in the variable declaration part.
\end{itemize}
A schema rule above is normally written as:
\[
a_1,\dots, a_n \to b,
\]
$x_1:C_1,\dots,x_m:C_m$, where $x_i, 1 \le i \le m$ are all
variables occurred in the schema rule.

Schema rules can be grounded into ground rules by assigning all
variables occurred in the schema rule to corresponding individuals.
In this case, the ground rule is called a {\em {ground} instance} of
the schema rule by the assignment. In this sense, a schema is
essentially a concept (i.e., set) of ground rules, containing all
its ground instances. Both ground rules and schema rules are called
{\em rules}. In particular, ground rules can be considered as schema
rules without variable declarations. Similarly, rules can be
considered as individuals.

\comment{ Schema rules are not new in the literature. In fact, many
existing production systems, e.g., context-free grammar, essentially
use schemas as well.}

A {\em structured production system} ({\em SPS} for short) is a pair
$\langle \mathcal{S}, \mathcal{R} \rangle$, where $\mathcal{S}$ is a
syntactic structure and $\mathcal{R}$ a set of (ground, schema)
rules such that all syntactic objects (including individuals,
concepts and operators) in $\mathcal{R}$ are defined in
$\mathcal{S}$.

\begin{example}\label{door-scenario-3}[Example \ref{door-scenario-2} continued]
According to the definitions, the following rule is applicable:
\small
\[
Status(door_1)=o, Do(Close(door_1)) \to Status(door_1)=c.
\]
\normalsize Note that this rule covers the transition not only from
$state_1$ to $state_2$ but also the one from $state_3$ to $state_4$.
Then, the door scenario with $n$ doors can be characterized by the
following four schema rules:
\begin{align*}
&Status(x)=o, Do(Close(x)) && \to && Status(x)=c, \\
&Status(x)=c, Do(Close(x)) && \to && Status(x)=c, \\
&Status(x)=o, Do(Open(x)) && \to && Status(x)=o, \\
&Status(x)=c, Do(Open(x)) && \to && Status(x)=o,
\end{align*}
where $x:Door$ is the variable declaration part.

It can be observed that using (schema) assertions and rules can be
much more succinct in comparison with state transition systems.
While the latter uses exponential number of symbols, we only need a
linear number of ground rules and a constant number of schema rules
to formalize this domain.
\end{example}

\section{Rule Application Strategies}

Rule application is a key issue in production systems. At a certain
stage, if the antecedent of a rule is satisfied, then this rule can
be triggered. It could be the case that many rules can be triggered
at the same time. However, only one rule can be applied. Then, the
consequent action will be executed or the consequent postcondition
can be obtained.

Nevertheless, in some cases, one may need different rule application
strategies. For instance, in cellular automata, the new status of
each cell is updated simultaneously based on the current statuses of
this cell itself and its neighborhood. Hence, naive rule application
strategy is not flexible enough.

In this section, we show that this issue can be addressed in
structured production systems by utilizing well-formed structures.
We first follow the same basic rule application strategy as
traditional production systems. Then, we show that, other different
rule application strategies, including simultaneous rule
application, constant rule and many more, can be reduced into the
basic one by utilizing syntactic structures.

\subsection{The basic strategy}

We start with the basic rule application strategy for structured
production systems. Similar to tradition production system, at each
stage, only up to one ground rule can be applied. Again, a ground
rule can be triggered if all the assertions in its antecedent are
true under the current state. Nevertheless, the ground rule can be a
genuine ground rule, or a ground instance of a schema rule with
corresponding assignment. If a ground rule is applied, then its
consequent $t_1=t_2$ needs to be satisfied by assigning the new
value of $t_1$ to be the existing value of $t_2$.

A {\em derivation} $d$ of an SPS is a sequence $r_1,\dots,r_n$ of
ground rules triggered and applied, denoted by $d=r_1; \dots; r_n$,
where ``$;$" is an operator connecting rules.

\subsection{Constant rules}

In some cases, one may wish some rules to be applied at all stages.
For instance, the following rule simply counts the global time clock
of an SPS.
\[
\to t=t+1,
\]
which means that after each stage, the counter $t$ is increased by
$1$. In order to make it work, this rule has to be applied at all
stages. We call them {\em constant rules}.

A constant rule can be reduced into the basic rule application
strategy by attaching it to all other rules in an SPS. For this
purpose, we introduce a special term structure called conditional
term. A {\em conditional term} is a triple $\langle \phi, t_1, t_2
\rangle$, where $\phi$ is an assertion and $t_1$ and $t_2$ two
terms. If the assertion $\phi$ holds, then this conditional term
equals to $t_1$; otherwise, it equals to $t_2$. By using conditional
term,s each production rule of the form
(\ref{simple-production-rule}), in which the consequent $b$ is
$t_1=t_2$, can be equivalently rewritten as
\begin{equation}\label{rule-rewritten-no-precondition}
\to t_1 = \langle a_1 \land \dots \land a_n, t_2, t_1\rangle.
\end{equation}

Let $r$ and $r'$ be two ground rules and $\widehat{r}=\to t_1=t_2$
and $\widehat{r'}=\to t'_1=t'_2$ their rewritten of the form
(\ref{rule-rewritten-no-precondition}) respectively. The rule
obtained from $r$ by {\em attaching} $r'$, denoted by $r\circ r'$,
is the following rule
\[
\to (t_1,t'_1)=(t_2, t'_2).
\]
Let $\mathcal{R}$ be a set of rules and $r'$ a rule. The rule base
obtained from $\mathcal{R}$ by {\em attaching} $r'$, is the set
$\{r\circ r' \D r \in \mathcal{R}\}$.

A {\em constant rule} in an SPS is a rule attached to all other
rules in the system.

\comment{In particular, we use the symbol $\Rightarrow$ to denote a
constant rule. That is, a constant rule is written as
\begin{equation}\label{form-constant-rule}
a_1, \dots, a_n \Rightarrow b,
\end{equation}
}

\subsection{Simultaneous rule application}

In some cases, one may want to apply some rules simultaneously. This
can be reduced to the basic rule application strategy by utilizing
syntactic structures as well. We use rule grouping for this purpose.
There are two different kinds of rule grouping, i.e., grouping a
finite set of ground rules and grouping a schema rule.

Grouping a finite set of ground rules can be achieved by rule
attaching as well. Let $r_i, 1 \le i \le n$ be a finite set of
ground rules. The {\em group rule} of $r_i, 1 \le i \le n$, also
denoted by $[r_1,\dots,r_n]$, is the following rule
\[
r_1 \circ r_2 \circ \dots \circ r_n.
\]
Once these rules are grouped together, they will be triggered and
applied simultaneously.

Grouping a finite set of ground rules yields a straightforward
extension of production rules to allow multiple assertions in the
consequents of rules.

Grouping a schema cannot simple be done by attaching as there could
be infinite number of ground instances of a schema rule. For this
purpose, we need to induce an ordering on sets. Let $S
=\{a_1,a_2,a_3,\dots,a_n,\dots\}$ be a countable set.\footnote{Here,
we only present the case that all concepts only contain countable
number of individuals.} By $\overrightarrow{S}$, we denoted the
following set $\{\{a_1\}, \{a_1,a_2\}, \{a_1,a_2,a_3\}, \dots,
\{a_1,a_2,a_3,\dots,a_n\},\dots\}$.

Let $r$ be a schema rule of the form $a_1,\dots, a_n \to t_1=t_2$
with variable declarations $x_i:C_i, 1 \le i \le m$. We first
rewrite it into $\to t_1=t$, where $t$ denotes the conditional term
$\langle a_1\land \dots\land a_n, t_2, t_1\rangle$. Note that both
$t_1$ and $t$ could contain variables. Essentially, $t_1=t$ means
that for all assignments $\eta=(x_1/d_1,\dots,x_m/d_m)$,
$t_1\eta=t\eta$. The {\em group rule} of $r$, denoted by $[r]$, is
the following rule
\[
\to \overrightarrow{\{t_1\eta\D \eta\textrm{ is an assignment}\}} =
\overrightarrow{\{t\eta \D \eta \textrm{ is an assignment}\}}.
\]
If the postcondition holds, then for all assignments $\eta$,
$t_1\eta=t\eta$, and vice versa.

\subsection{Preference over rules}

In some cases, one may wish a rule is more preferred than another.
That is, if the former rule is applicable, then always trigger and
apply it. Otherwise, one can check whether the latter rule is
applicable or not.

Let $r$ and $r'$ be two rules, and $r$ is more preferred than $r'$,
written by $r \succ r'$. In order to simulate this preference
relationship, we introduce an operator $Applicable$ over rules.
$Applicable(r)$ means that the preconditions of $r$ are all
satisfied so that rule $r$ can be triggered and applied. Then, we
add a new precondition to rule $r'$, stating that $r'$ is applicable
only if rule $r$ is not applicable at the moment, that is,
$Applicable(r)$ has to be false.

Formally, let $r$ be a ground rule and $a_1,\dots,a_n$ all its
preconditions. Let $r'$ be a ground rule, $a'_1,\dots,a'_n$ all its
preconditions and $b'$ its postcondition. To capture the preference
relationship $r \succ r'$, we group the following rules together,
including rule $r$, rules of the form
\[
\lnot a_i \to Applicable(r)=false,
\]
where $1 \le i \le n$, and
\[
a'_1,\dots,a'_n, Applicable(r)=false \to b'.
\]

This is a finite group of rules, which means that these rules will
be triggered and applied simultaneously. Rule $r$ is in the group,
meaning that $r$ can be triggered and applied in any circumstance if
its preconditions are satisfied. The rules $a_i \to
Applicable(r)=false$ mean that if one of the preconditions of $r$ is
not satisfied, then rule $r$ is not applicable. Finally, the rule
$a'_1,\dots,a'_n, Applicable(r)=false \to b'$ means that the rule
$r'$ can be applied only if rule $r$ is not applicable. In this
sense, rule $r$ is always more preferred than rule $r'$.

To extend this for schema rules, one needs to deal with the
assignments, which can be included in the scope of the newly
introduced operator $Applicable$.

\subsection{Ordered rule application}

Sometimes one may wish the rules to be applied in an order, i.e., a
rule can be applied only if another rule is already applied. A
special case is sequential rule application, i.e., rules are applied
one by one. We show that ordered rule application and sequential
rule application can be reduced into the basic rule application
strategy as well.

We introduce a new operator $Applied$ over all rules, explicitly
monitoring whether a rule is applied or not. Let $r$ be a ground
rule of the form (\ref{simple-production-rule}) and $r'$ is the rule
that has to the applied before the application of $r$, denoted by
$r'\rhd r$. We rewrite $r'$ as
\[
a'_1,\dots,a'_n \to b', Applied(r').
\]
and $r$ as
\[
a_1,\dots, a_n, Applied(r') \to b, Applied (r).
\]
According to the construction, this rule can be triggered only if
rule $r'$ is applied. After applying this rule, rule $r$ is set to
be applied. Hence, rule $r$ can only be applied after the
application of $r'$. Ordered rule application on schema rules can be
done similarly except that one needs to deal with assignments, which
can be included in the scope of the operator $Applied$. Sequential
rule application is a special case of ordered rule application when
a total order is enforced on all rules.

\section{Capturing Existing Approaches}

In this section, we argue that structured production system provides
a general framework for modeling dynamics and automated reasoning by
showing that it can capture many existing approaches.

\noindent{\bf Traditional production system} Clearly, traditional
production systems are special cases of structured production
systems. One issue in traditional production system is that the
consequent could be either an action or a post-condition. In
structured production systems, we can unify them together by
introducing a special operator $Do$ on all actions to convert them
into assertions, as shown in Example \ref{door-scenario-1}. Although
structures are used in some traditional production systems, their
power are not thoroughly investigated. In this extended abstract, we
further show that the richness of syntactic structures can indeed
bring a lot of benefits to production systems.

\noindent{\bf Subsumption architecture} Subsumption architecture
\cite{Brooks85} is an extension of traditional production system by
allowing multi-layer of production rules to be applied
simultaneously, where lower-level actions are sub-behaviors of
higher-level ones. Subsumption architecture can be considered a
special case of structured production systems as well in the sense
that it utilizes a syntactic structure to model the hierarchical
relationships among actions. Also, parallelism can be implemented in
SPS by rule grouping.

\noindent{\bf Automata and Turing machines} As the foundation of
computational theory, automata and Turing machine play a critical
role in computer science. An automaton (such as a Turing machine)
can be reformulated as a structured production system, where each
item in the transition function forms a production rule and the rest
(including states and symbols) is defined by a well-formed
structure.

\noindent{\bf Abstract rewriting systems and state transition
systems} Abstract rewriting systems and state transition systems are
simple models for modeling dynamics.Similar to automaton, an
abstract rewriting system or a state transition system can be
re-formulated as a SPS, again, in which the transitions are modeled
by production rules and the rest is captured by a structure.

\noindent{\bf Axiom systems} Logic axiom systems are often
considered as deliberative that are very different from production
systems. Interestingly, logic axiom systems can be converted into
structured production systems as well. For this purpose, we
introduce an operator $Prove$ operating on all well defined formulas
whose value can be either true or false. Then, axioms and inference
rules in logic axiom systems can be translated into schema
production rules with variables ranging over all well-defined
formulas. For instance, the exclusive middle axiom
\[
P \lor \lnot P
\]
is translated into a schema rule
\[
\to Prove(P \lor \lnot P),
\]
with the variable declaration $P:\mathcal{L}$, where $\mathcal{L}$
is the language (i.e., a concept) of all well-defined formulas.
Similarly, the Modus Ponens inference rule is translated into
\[
Prove(P), Prove(P \supset Q) \to Prove(Q),
\]
where $P,Q:\mathcal{L}$.

\noindent{\bf Cellular automata} A cellular automaton
\cite{Neumann66,Schiff07} consists of a grid of cells whose values
range over a finite set. At each stage, the new value of each cell
is only depending on its adjacent cells (called the neighborhood) by
some fixed rules. Clearly, a cellular automaton can be regarded as a
SPS in the sense that the rules governing the value change of cells
can be straightforwardly converted into a production rule, while the
rest, including the grid itself, can be captured by a syntactic
structure. No matter the rules are applied synchronously or
asynchronously, this can be captured in structured production
systems with different rule application strategies discussed in
Section 4.

It can be seen that many other approaches, e.g., opinion dynamics
\cite{Bindel-etal15,HegselmannK02} and, can be reformulated as
structured production system as well.

\comment{ Opinion dynamics \cite{Bindel-etal15,HegselmannK02} is
similar to cellular automata but under a different context, normally
in a social network. Each node in a social network has a certain
opinion on some topic. At each stage, each node's opinion is
affected and then updated based on its own and its neighbors'
opinions by some fixed rules, e.g., majority voting. Similar to
cellular automata, opinion dynamics can be considered as structured
production systems as well.}

\section{Handling Uncertainty}

One of the main concerns of traditional production systems is that
production rule of the form (\ref{simple-production-rule}) is too
simple to model sophisticated application domains, for instance, to
handle uncertainty. In this section, we show that this issue can be
addressed by utilizing the syntactic structures.

\subsection{Uncertainty associated with assertions}
One way to incorporate uncertainty in structured production systems
is to extend assertions with uncertainty information. For instance,
let $\phi$ be an assertion and $Pr$ a probability function whose
domain is the concept of all assertions and whose value is a real
number between $0$ and $1$. Then, $Pr(\phi)$ is an individual.
Consequently, $Pr(\phi)=0.6$ is a probabilistic assertion, meaning
that the probability of $\phi$ to be true is $0.6$. It is easy to
see that other uncertainty assertions such as fuzzy assertion can be
defined in a similar way.

With uncertainty assertions, one can directly talk about uncertainty
in structured production systems. For instance, the following schema
rule
\begin{align*}
Pr(Smoke(x))=0.9, Pr(Cancer(Father(x)))=0.85 \\ \to Pr(Cancer(x))=
0.045,
\end{align*}
where $x:Human$, means that if the probability of a person $x$ being
a smoker is $0.9$ and the probability of $x$'s father having a
cancer is $0.85$, then the probability of $x$ getting a cancer is
$0.45$. The syntactic objects in the rule could vary or could be
more abstract. For instance,
\begin{align*}
Pr(Smoke(x)=a), Pr(Cancer(Father(x))=b)  \\ \to Pr(Cancer(x))=
f(a,b),
\end{align*}
is a more abstract schema rule for this scenario, where $f$ is an
arithmetic function.

Similar to handling logic axioms and inference rules, one can encode
some theorems and axioms about uncertainty, e.g., Kolmogorov's
probability axioms, by structured production rules. As an example,
Bayes' theorem can be encoded into the following schema rule:
\begin{equation}\label{Bayes-theorem}
{\to \displaystyle Pr(A\mid B)={\frac {Pr(B\mid A)\,Pr(A)}{Pr(B)}},}
\end{equation}
where $A$ and $B$ range over all assertions and $\mid$ is an
operator for conditional assertions.

\subsection{Uncertainty associated with rules}
An alternative way for handling uncertainty in structured production
systems is to attach uncertainty information to rules. For instance,
let $r$ be a rule. We introduce a probability function $Pr$ whose
domain is the concept of all rules and whose value is a real number
between $0$ and $1$. Then, $Pr(r)$ is an individual, and
consequently, $Pr(r)=0.8$ is a probabilistic assertion, meaning that
the probability of $r$ to be true is $0.8$.

One can extend this to a probability function over derivations. For
instance, we can define the following schema rule to calculate the
probabilities of derivations
\begin{equation}\label{probability-product}
d= r_1; \dots; r_n  \to Pr(d) = Pr(r_1)\times \dots \times Pr(r_n)
\end{equation}
where $d$ ranges over all derivations and $r_i, 1\le i \le n$ ranges
over all rules.

\subsection{Embedding probabilistic context-free grammar}

Following the above ideas of handling uncertainty in structured
production system, one can see that many interesting approaches, for
instance, probabilistic context-free grammar that is widely used in
natural language processing \cite{lari90}, can be considered as
structured production systems as well.

Formally, a probabilistic context-free grammar is a quintuple
$\langle M,T,R,S,P \rangle$, where $M$ is a set of intermediate
symbols including the start symbol $S$, $T$ a set of terminal
symbols disjoint from $M$, $R$ a set of rules of the form
\begin{equation}\label{cfg-rule}
A \to \alpha,
\end{equation}
where $A \in M$ and $\alpha$ a string of symbols over $M \cup T$,
and finally, $P$ a probabilistic function from $R$ to $[0,1]$. A
derivation is a sequence of rule applications, generating a string
of terminal symbols from the start symbol $S$. At the beginning, the
string is merely the start symbol $S$. Then, at each stage, in order
to obtain the next string, one picks up a rule of the form
(\ref{cfg-rule}) such that $A$ is in the current string and replace
$A$ with $\alpha$. One repeats the above process until the string
only consists of terminal symbols. Finally, the probability of the
derivation is the product of the probabilities of rules used in
every stage.

To re-formulate probabilistic context-free grammars in structured
production systems, we need to define a syntactic structure. We
borrow the concepts, including $M$ and $T$, defined in the grammar.
Specifically, $S$ an individual. We use $String$ to denote a concept
of all possible strings over $M\cup T$, and $\bullet$ the
concatenation operator over strings. We also introduce a concept $D$
of all derivations such that $R \subseteq D$, and $;$ the operator
that connects two derivations. Finally, we specifically introduce an
individual $cs$ to denote the current string whose initial value is
$S$, and $cd$ to denote the current derivation whose initial value
is empty.

Then, we translate a rule $r \in R$ of the form (\ref{cfg-rule})
into the following schema production rule:
\[
cs=s\bullet A \bullet s' \to cs=s \bullet \alpha \bullet s',
cd=cd;r,
\]
where $s, s':String$. Together with the schema rule
(\ref{probability-product}) to calculate the probabilities of
derivations, a probabilistic context-free grammar is converted into
a structured production system.

To end up with this section, it is worth mentioning that one can
combine these two different ways of handling uncertainty together.
It is valid to state the following rule
\begin{align*}
& Pr(Cancer(Father(x)))=0.7 \\ \to & Pr(Smoke(x) \to
Pr(Cancer(x)=0.02))= 0.8,
\end{align*}
where $x:Human$.

\section{Conclusions, Discussions and Future Work}

In this extended abstract, we proposed structured production systems
that enhance tradition production systems with well-formed syntactic
structures. For using structured production systems to model an
application domain, objects in the domain are represented by
individuals, concepts and operators; knowledge are formalized by
assertions; finally, dynamics in the domain are captured by (schema)
production rules.

Production system is one of the most important AI approaches.
Nevertheless, it has been less studied in the AI community in recent
decades. Perhaps one reason is that it is considered to be too
simple. We argue that simplicity should never be an issue in
scientific research. On the contrary, following the Occam's razor
principle, the simpler, the better. This is the case especially for
production system, providing its tremendous applications in AI
including expert systems, natural language processing and robotics.

Although simple, research on production system is far from mature.
There are several critical issues with traditional production
systems, including succinctness, expressiveness and flexibility. In
this extended abstract, we showed that these issues can be addressed
by introducing well-formed syntactic structures. Due to the richness
of structures, we showed that structured production systems are
expressive enough to model sophisticated application domain in a
succinct and flexible way. As an evidence, we showed that structured
production systems can handle uncertainty information, capture
different rule application strategies and many fundamental
approaches in computer science.

Another critical issue of production systems is the knowledge
acquisition problem, that is, how to engineer the production rules
at the first place. Although not discussed in this extended
abstract, it is another important motivation of our work on
structured production systems. We plan to use well-structured
production rules themselves to generate and learn (schema)
production rules. We leave this as one of our most important future
works.

\comment{ For instance, the following schema rule specifies a
general induction principle to learn a schema assertion whether a
concept $C$ is a subclass of another concept $D$:
\[
c_1 \in C \land c_1 \in D, \dots, c_n \in C \land c_n \in D,
\not\exists (c \in C) \land (c \not \in D) \to C \subseteq D ,
\]
where $C$ and $D$ range over some pre-defined concepts, $c_1, \dots,
c_n$ are some individuals where $n$ is a predefined number.
Intuitively, this rule means that, if we observe that many
individuals in $C$ are also in $D$ and there is no counterexample,
then we conclude that $C$ is a subclass of $D$. Production rule
learning by production rules themselves is a critical problem for
structured production systems. }

Structured production systems are nondeterministic. Firstly, similar
to traditional production systems, there could be many different
rules applicable at a certain stage. The system needs to determine
which one to trigger and apply. Secondly, schema rules may have
different groundings, which leads to nondeterminism as well. To
address this issue, traditional production systems often have a rule
matching mechanism as well as a rule selection mechanism.
Nevertheless, in structured production system, we plan to use an
alternative approach that encodes rule matching and selection
themselves as production rules with the help of syntactic
structures. We leave this as another future direction.

Both production systems and structures are not new in the
literature, neither is their integration. In fact, many existing
production systems, such as context-free grammars, already use
schema rules. Nevertheless, most of them only consider their
integrations by schema rules. The main contribution of this work is
to further advocate their marriage to show that structures can
indeed bring much more benefits to production systems. In this
extended abstract, we showed that structures can be used not only in
schema rules but also in a more flexible way to address many key
issues in production systems.

As mentioned in the introduction, one of our intended application
domains is intelligence tests including sequencing number games. For
such a challenging task, we need not only to represent complicated
syntactic objects including different patterns and their
combinations but also to effectively reason about them. We use
well-formed structures for the former while production rule based
reasoning for the latter. This extended abstract is focused on the
theoretical part, and we will present our preliminary results on
sequencing number games based on structured production systems in
another paper.

\comment{undecidable, soundness and completeness}

Finally, we argue that structured production system has the
potential to bypass the long standing curse of symbolic AI that
always tries to find a balance between expressiveness and
efficiency. In traditional symbolic AI, a critical dilemma is the
tradeoff between expressiveness and efficiency (often measured by
computational complexity) \cite{LevesqueB87}. The more expressive
power a symbolic AI formalism has, the less efficiency it is, or the
other way around. Then, many researches are devoted into adding or
removing some building blocks in symbolic AI formalisms in order to
make a good balance between them. Nevertheless, in many application
domains, both expressiveness and efficiency are highly needed. We
argue that structured production system suggests a promising
solution for solving this dilemma. While expressiveness for
representation is achieved by syntactic structures, efficiency for
reasoning can be obtained by applying production rules.

\section*{Acknowledgement}

The author would like to thank Prof. Fangzhen Lin for his comments
on a first draft of this paper.

\bibliographystyle{plain}
\bibliography{ijcai17}

\end{document}